\documentclass[10pt,twocolumn,letterpaper]{article}

\usepackage{wacv}
\usepackage{times}
\usepackage{epsfig}
\usepackage{graphicx,subfigure}
\usepackage{amsmath,amssymb,url,verbatim,indentfirst,afterpage,color}

\usepackage[pagebackref=true,breaklinks=true,letterpaper=true,colorlinks,bookmarks=false]{hyperref}

\wacvfinalcopy % *** Uncomment this line for the final submission

\def\wacvPaperID{386} % *** Enter the wacv Paper ID here

\ifwacvfinal\fi
\begin{document}

%%%%%%%%% TITLE
\title{Automatic Discovery and Geotagging of Objects from Street View Imagery}

\author{Vladimir A. Krylov\qquad Eamonn Kenny\qquad Rozenn Dahyot\\\\
%School of Computer Science and Statistics, Trinity College Dublin, Ireland\\
ADAPT Centre, School of Computer Science and Statistics, Trinity College Dublin, Dublin, Ireland\\
%{\tt\small vladimir.krylov@adaptcentre.ie}
}

\maketitle
\linepenalty=1000

%%%%%%%%%%%%%%%%
\begin{abstract}

Many applications such as autonomous navigation, urban planning and asset monitoring, rely on the availability of accurate information about objects and their geolocations. In this paper we propose to automatically detect and compute the GPS coordinates of recurring stationary objects of interest using street view imagery. Our processing pipeline relies on two fully convolutional neural networks: the first segments objects in the images while the second estimates their distance from the camera. To geolocate all the detected objects coherently we propose a novel custom Markov Random Field model to perform objects triangulation. The novelty of the resulting pipeline is the combined use of monocular depth estimation and triangulation to enable automatic mapping of complex scenes with multiple visually similar objects of interest. We validate experimentally the effectiveness of our approach on two object classes: traffic lights and telegraph poles. The experiments report high object recall rates and GPS accuracy within 2 meters, which is comparable with the precision of single-frequency GPS receivers.

%Many applications rely on the availability of information about objects and their accurate locations including autonomous navigation, urban planning, asset monitoring, to name a few. In this study we address automatic detection and geotagging of object classes based on street view imagery. To perform object detection we develop and train deep fully Convolutional Neural Networks (CNNs) on specific object types and propose a novel optimization procedure for geotagging the detected object-instances based on triangulation and depth mapping. The experimental study performed on two object classes - traffic lights and telegraph poles - demonstrates high precision in both identifying and geotagging object-instances from Google Street View imagery under mild object sparsity assumption. The experiments performed in dense urban and rural environments reveal the high potential of the proposed automatic image processing pipeline for asset inventory problems.

\end{abstract}

%%%%%%%%%%%%%%%%
\section{Introduction}

The rapid development of computer vision and machine learning techniques in recent decades has excited the ever-growing interest in automatic analysis of huge image datasets accumulated by companies and individual users all around the world.
Image databases with GPS information, such as Google Street View (GSV) and images posted on social networks such as Twitter, are now widely available online and can be queried seamlessly using APIs, conveniently set up and regularly updated with new image data by the providing companies. 
Street view imagery (e.g., from GSV, Bing Streetside, Mapillary) represents a collection of billions of geotagged images covering millions of kilometers of roads and depicting street view scenes collected at regular intervals. This incredible amount of visual data allows one to address a multitude of detection and mapping problems by exploring areas remotely through imagery. 

%Such tasks require careful design of automated algorithms that could efficiently exploit data from these vast image repositories.

\begin{figure}[t!]
\begin{center}
\includegraphics[width = \linewidth]{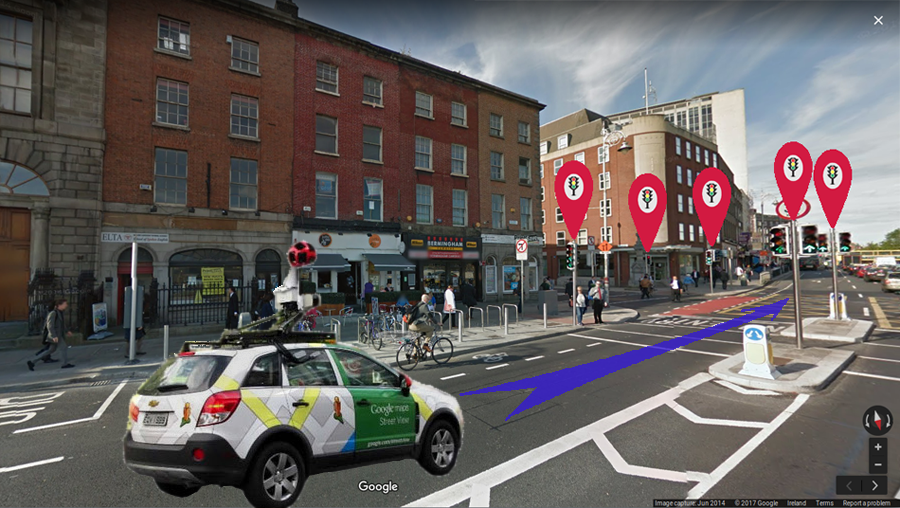}
\end{center}
\vspace{-0.8\baselineskip}
\caption{Detection and geotagging of stationary objects of interest (traffic lights) from Google Street View imagery.}
\label{figIntro}
\end{figure}

A lot of research has been dedicated to leveraging street view imagery in combination with other data-sources such as remotely sensed imagery~\cite{mattyus2016hd,wegner2016cataloging} or crowd-sourced information~\cite{hara2015improving} to discover particular types of objects or areas.
Here we address the problem of automated discovery and geotagging of recurring objects using street view images as a sole source of input data, see Fig.~\ref{figIntro}.
We consider any class of stationary objects sufficiently compact to have a single geotag that are typically located along the roads, like street furniture (post boxes, various poles and street-lamps, traffic lights and signs, transport stops, benches, etc.), small facade elements (cameras, antennas, security alarm boxes, etc.) and minor landmarks.  
Inventory and precise GPS mapping of such objects is a highly relevant task and, indeed, OpenStreetMap and Mapillary are currently encouraging their users to contribute such information to their databases manually.
Nevertheless, the vast majority of these objects can be mapped automatically by efficiently exploring the publicly available street view imagery.
To the best of our knowledge, no solution to this problem relying on street view imagery alone has been proposed in the literature.

\begin{figure*}[t!]
\begin{center}
\includegraphics[width = \linewidth]{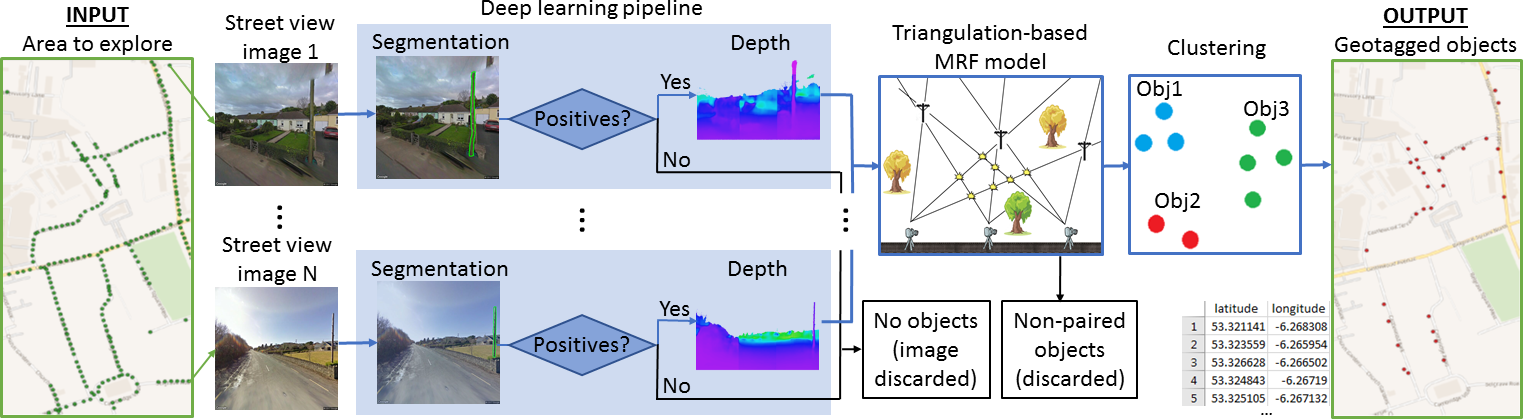}
\end{center}
\vspace{-1.2\baselineskip}
\caption{Proposed geotagging pipeline: from an area of interest with street view images (green dots) to geotagged objects (red dots).}
\label{figflow}
\end{figure*}

We propose a complete image processing pipeline for geotagging of recurring objects from street view imagery. The main components in the pipeline, see Fig.~\ref{figflow}, are: two state-of-the-art fully convolutional neural networks (FCNN) for semantic segmentation and monocular depth estimation, and a novel geotagging model that geolocates objects by combining depth information and geometric triangulation through a Markov Random Field (MRF) formulation.

In this work we address geotagging of recurring stationary objects that may be partially or completely occluded in some of the input images. We do not resort to any explicit geometrical modeling of object shapes, nor do we rely on any object location patterns.
In order to enable automatic geolocation we enforce a mild assumption of object sparsity: we expect the objects to be located at least 1m apart.
We evaluate the performance of the proposed pipeline on two object classes: traffic lights and telegraph poles. We perform extensive experimental analysis to confirm the accuracy of both object discovery and geotagging. 

{The central contribution of this work is the proposed MRF-based geotagging procedure. Unlike previous works~\cite{wacv17,iccv11}, this technique allows us to leverage monocular depth estimates to automatically resolve complex scenes containing multiple instances of identical objects. The proposed pipeline is modular which makes it possible to replace segmentation and depth modules with alternative techniques or pretrained solutions for particular object families.}

The paper is organized as follows:
we first review some relevant state-of-the-art approaches in Sec.~\ref{sec:prior}. Our complete geotagging pipeline is presented in  Sec.~\ref{sec:our:approach} and then validated experimentally in Sec.~\ref{sec:exps}. Sec.~\ref{sec:concl} concludes this study.

%%%%%%%%%%%%%%%%
\section{Related work}
\label{sec:prior}

In the last decade a considerable effort has been directed towards intelligent use of street view imagery in multiple applied areas: mapping~\cite{hara2015improving,wegner2016cataloging}, image referencing~\cite{hays2008im2gps}, navigation~\cite{mattyus2016hd}, rendering and visualization~\cite{Bulbul201728,Du2016SocialStreetView}, etc. 
Methods designed in~\cite{mattyus2016hd,wegner2016cataloging} use street view  combined with aerial imagery to achieve fine-grained road segmentation, and object detection (trees), respectively. These methods rely on object discovery through street view imagery and recover position from aerial data. In~\cite{hays2008im2gps} street view images are used as reference to geolocate query images by resorting to scene matching. The GSV imagery is employed in~\cite{Bulbul201728,Du2016SocialStreetView} in conjunction with social media, like Twitter, to perform visualization and 3D rendering. 

{Several methods have been developed to map particular types of objects from street-level imagery: manholes~\cite{iccv11}, telecom assets~\cite{wacv17}, road signs~\cite{soheilian2013detection}. These methods rely on triangulation from individual camera views to geolocate the considered road furniture elements. All three approaches rely heavily on the geometrical and visual clues to perform matching whenever multiple objects are present in the same scene. 
%Consequently, if multiple identical objects of interest are present in a scene these methods perform poorly.
As a consequence, whenever multiple identical objects are present simultaneously, these methods perform poorly. In particular, as pointed out in~\cite{wacv17}, visual matching performed via SIFT on street-level images with strong view-point position change has limited performance in establishing reliable matching between small objects.
}
%In~\cite{soheilian2013detection} geotagging of road signs is performed based on explicit geometrical shape modeling, on images obtained by a custom street view camera.

\begin{comment}
In this work we are interested in geolocating specific objects of interest in the 3D environment and to this end we propose to use a combination of computer vision techniques, in particular semantic segmentation (reviewed in section \ref{sec:PR}) and monocular depth estimation approaches (reviewed in section \ref{sec:DE}).

combining test time information from multiple views of each geographic location (e.g., aerial and street views), dataset: Pasadena Urban Trees
~\cite{wegner2016cataloging}

joint inference over both, monocular
aerial imagery, as well as ground images taken from a
stereo camera pair mounted on top of a car~\cite{mattyus2016hd}

method  for  collecting  bus  stop location  and  landmark  descriptions  using  online crowdsourcing  and  Google  Street View 
~\cite{hara2015improving}

estimating a distribution over geographic locations from a single image using a purely data-driven scene matching approach
~\cite{hays2008im2gps}

\subsection{Object detection}
\label{sec:PR}
\end{comment}

Image segmentation is one of the central tasks in object extraction.
A multitude of approaches have been proposed to address this problem: starting from  elementary pattern analysis techniques, such as Hough transform and local gradients, through feature extraction-based tools that rely on cascades of weak classifiers~\cite{ViolaJones2001}, to more advanced machine learning methods such as random forests, support vector machines and convolutional neural networks (CNNs)~\cite{resnet,vgg}. The latter have recently pushed the machine vision techniques to new heights by allowing automatic feature selection and unprecedented capacity to efficiently learn from huge volumes of data. FCNNs are a natural form of CNNs when addressing dense segmentation since they allow location information to be retained through the decision process by relying solely on convolutional layers~\cite{shelhamer2017fully}. Their output can be in the form of bounding boxes~\cite{ren2017faster,liu2016ssd} or segmentation maps. The latter are obtained via deconvolutions only~\cite{shelhamer2017fully} or by resorting to conditional random fields~\cite{crfasrnn}.

\begin{comment}
Many approaches to pattern detection in images have been proposed in the past decades.
Early techniques such as the Hough Transform in the early 80's, focused on elementary patterns such as lines.
In the early 00's, machine learning techniques proposed to learn well chosen features extracted from a set of exemplars  using a combination of weak classifiers, and this has led to major improvements in applications such as face detection  for instance \cite{ViolaJones2001}. OpenCV library \cite{opencv_library} has popularized many such image processing algorithms as well as  providing an easy pipeline for training a new class of objects to detect in images.
More recently  deep learning has pushed the performance to higher ground by also learning what features are relevant to use for detecting the patterns in image data leveraging the need of expert knowledge in image processing and computer vision for building a pattern detection algorithms. Many libraries are now readily available for using deep learning as part of an image processing pipeline.

\subsection{Scene analysis and depth estimation}
\label{sec:DE}
\end{comment}

Estimation of camera-to-object distances and, more generally, 3D scene analysis from RGB images, can be addressed in several ways. The most explored are the stereo-vision approaches~\cite{seitz2006comparison} that estimate camera-to-object distances from multi-view stereo image disparity analysis or perform scene reconstruction via Structure-from-Motion methods~\cite{sfm}. These rely on various assumptions about camera positions and trajectory, and typically require a rich set of input RGB images and a certain form of knowledge about the analyzed scene. These classes of methods rely heavily on feature extraction and matching. Another way to address scene depth evaluation has been explored in several recent studies~\cite{GodardAB16,laina2016deeper,li2015depth}. The central idea in these is to recover depth information from monocular images based on the extensive scene interpretation capabilities of FCNNs by training them on RGB and depth or disparity images. These methods achieve consistent yet approximate results relying solely on the information available in the color bands.
%relative object positioning and perspective information available in the input color bands. 

%%%%%%%%%%%%%%%%
\section{Object discovery and geotagging}
\label{sec:our:approach}

Street view images are harvested by defining the area of interest with the corresponding GPS coordinates and querying the API to download the relevant geotagged imagery.
%The latter is provided along with camera GPS locations and headings. 
Each image is processed independently allowing efficient parallelization of this most time-consuming step. All images with discovered objects, as reported by the segmentation FCNN, are processed by the depth evaluation FCNN to evaluate camera-to-object distances, see Fig.~\ref{figflow}.
Image processing results are then fed into a MRF model and clustered in order to obtain a coherent list of triangulated objects.

\subsection{Object  segmentation}
One of the best-performing state-of-the-art semantic segmentation FCNN models~\cite{shelhamer2017fully} is used in our pipeline as it conveniently outputs pixel-level labels that can be used directly as a mask for the depth estimation step. 

The subsampling part of this segmentation FCNN architecture is that of VGG-16~\cite{vgg} with the fully-connected layers converted to convolutional layers. These are followed by the upsampling (deconvolution) part that reverts from low-resolution segmentation to the original spatial resolution, in several steps. The steps in this procedure allow the combination of coarse segmentation maps with appearance information from shallower fine layers which have been generated by the subsampling part of FCNN. We employ the FCN-8s all-at-once model that reports the finest segmentation details by performing deconvolutions in the form of bilinear interpolations in three consecutive steps. 

If we consider a single class of objects we reshape all upsampling layers as well as the bottom convolution in the subsampling part to have two outputs - one for background and another for the objects of interest. It is also possible to perform detection of several object classes simultaneously by further increasing the number of outputs. 
%In the following we will consider a single class of objects.

We further add a second loss function (with the same weight) that penalizes only the false positives, thus effectively re-weighting the total loss to penalize false positives with a higher weight. This is due to our interpretation of the results where a small group of false positive pixels may result in a false positive object discovery, whereas missed pixels of a true object (false negatives) are less critical.

\subsection{Monocular depth estimation}
\label{sec:depth:estimation}
It is challenging to resort to stereo-vision approaches operating on street view imagery collected with the average acquisition step of 5 to 10 meters. The main reason is that such sequences are typically characterized by substantial mismatch in scenes due to new objects on road sides, object scaling, moving vehicles, occlusions, panorama stitching artifacts and, occasionally, distinct image acquisition dates. Furthermore, no a priori information is available about the contents of analyzed scenes and their geometry. Thus, we choose to rely on approximate estimates reported by a state-of-the-art FCNN depth estimation pipeline introduced in~\cite{laina2016deeper}. This architecture is composed of a fully-convolutional version of ResNet-50~\cite{resnet} followed by a cascade of residual up-projection blocks to obtain a dense depth map at the native image resolution. This pipeline is employed with no modifications and is fed only with the images where objects of interest are discovered. A unique depth estimate for each object segment is obtained by taking the $10\%$-trimmed mean of the depths of its constituent pixels. We apply trimming to gain robustness with respect to segmentation errors, in particular along the object borders.

%%%%%%%%%%%%%%%%%%            
\subsection{Object geotagging}
\label{sec:GPStagging}

\subsubsection{Location from single view}

Consider a single street view image where objects of interest have been segmented. For each segmented region, a ray can be traced in the 3D space going from camera towards the barycenter of the segmented region. A depth estimate for that region, converted via averaging into a single object-level distance, allows the object's GPS location to be estimated.    
%In order to recover the exact position of an object relative to a camera position one needs to accurately evaluate the distance to this object. Coupled with its relative orientation retrieved from the segmented image, this allows to automatically recover the object's geotag. 
Specifically, from the segmentation map we extract the geo-orientation at which the object is located relative to the camera, shift the position by the estimated distance, and finally convert the metric coordinates into the employed projection system (e.g., WGS-84). Our operational assumption is that the objects are sufficiently sparse to be successfully identified as separate instances by the segmentation pipeline. Hence, whenever multiple objects are tightly clustered (less than 1 meter apart) they are recovered as a single object instance with a unique geotag.

There are two limitations associated with this strategy: first the depth estimates are not very accurate, and second, the list of all objects detected in all input images is largely redundant due to objects observed from multiple camera positions. Indeed, each of such objects spawns multiple detected object instances with distinct GPS coordinates due to depth evaluation inaccuracies. 
%Thus, as an alternative to direct depth-based geotagging, we consider strategies where each object is observable from several camera positions. 

\subsubsection{Location from multiple views}

When an object is observed multiple times the image segmentation map needs to be matched to group together observations of the same object-instances. 
%Such matching could also help estimate a more accurate GPS location using standard strategies from multiple view geometry. 
%However our objects of interest (e.g. poles) are not distinguishable if more than one instance actually occurs  in the scene, and good correspondences are almost impossible to find on such almost texture-less object.  
Matching could be addressed in two ways: either by relying on general image features, such as SIFT, HOG, etc., or custom features extracted by the segmentation CNN. We have explored both ways but neither provides satisfactory quality of matching due to the strong degree of similarity between the objects in the considered object-classes. 
%This results in very low general intra-class feature variance. 
Specifically, street-side objects tend to be identical in the same area, e.g. a line of poles along a road or traffic lights around the same road junction. For this reason, the main source of dissimilarity in image matching is provided by the background and / or occlusions in the object bounding boxes. The background composition, however, is subject to changes with the rotation of the viewing angle. 
{In particular, similarly to~\cite{wacv17}, we have observed that visual matching is not sufficient to reliably identify individual object instances in scenes with multiple similar or identical objects of interest.}
%Note also that the considered objects are in the proximity of roads, hence a small change in angle may result in strong dissimilarity of image background and thus upset object matching. 
%Consequently, we explore alternatives to achieve object-instance pairing.

\begin{figure}[t!]
\begin{center}
\includegraphics[width = \linewidth]{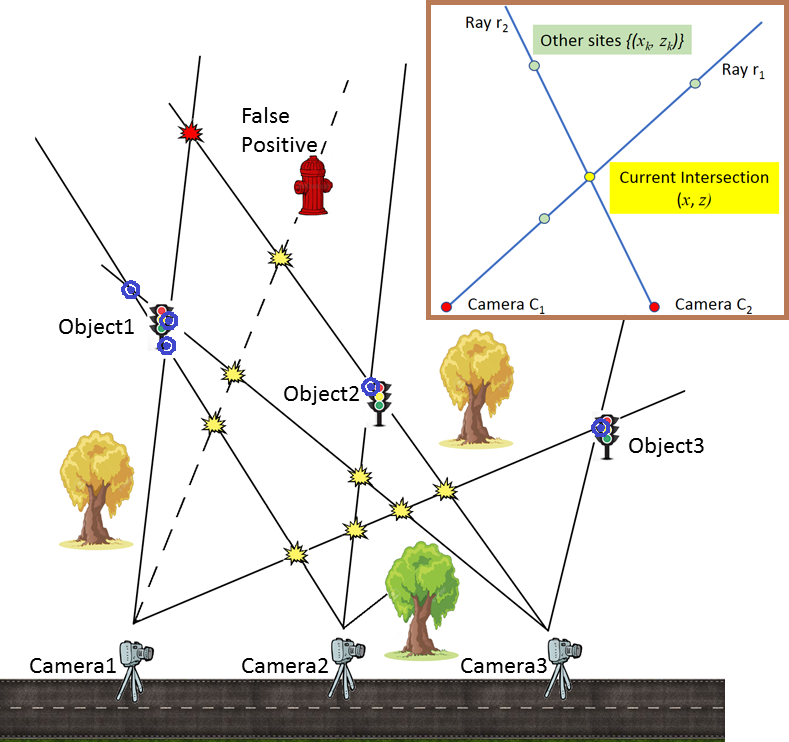}
\end{center}
\vspace{-0.5\baselineskip}
\caption{Object positions are identified based on triangulation of view-rays from camera positions. An example of irregular neighborhood of nodes is demonstrated in the side panel.}
\label{figIntersect}
\end{figure}

To perform geotagging we rely on a mixed approach based on triangulation. 
For illustration purposes in Fig.~\ref{figIntersect} we present an example of a simple stationary scene observed by three cameras, or equivalently, by the same camera at three distinct time moments. Three objects of interest (traffic lights) are present in the scene. Object1 is detected in all images, Object2 is missed in image collected by Camera1 (segmentation false negative), and Object3 is occluded from Camera2. A false positive is picked on Camera1 image. Fig.~\ref{figIntersect} demonstrates the complexity of accurately triangulating three objects in the space of 13 intersections. Note that the possible intersections are located such that clustering or grouping them without interpretation brings to wrong object maps.

{To address the problem of automatic mapping in multi-object scenes we propose an MRF-based optimization approach. This flexible technique allows us to incorporate the approximate knowledge about the camera-to-object distances into the decision process. Furthermore, it can be seen as generalization to the previously proposed methods~\cite{wacv17,iccv11} as it allows the integration of geometry/visual-based matching as additional energy terms. The designed MRF model operates on irregular grid and features custom energy terms appropriate for the considered triangulation problem.}

%The latter is used to define the initial space of view-ray intersections and their corresponding GPS coordinates. In the most simple scenario, when a single object of interest is observed and successfully identified on two distinct camera positions, the triangulation results in a single possible location for the object. If a second object is present and both are identified on two images the matching of object instances becomes necessary. As mentioned above the image-based matching has unsatisfactory performance. Hence, we employ depth map estimation to guide the instance matching. Specifically, we consider all possible object positions (two configurations in case of two objects observed by two cameras), estimate corresponding object-camera distances based on triangulation and accept the configuration that matches best the distances evaluated based on monocular image depth evaluation. An object-segment from a segmentation map contributes to a final geotagged object only when a suitable pairing with a segment from another image is found. Notice that to find a suitable match for triangulation we consider all objects (segments) in images available within 25m from the current camera position. Hence, any segment can be paired with multiple other segments: with up to one from any other images. If no suitable pairing is found for an object, it is rejected as a false positive committed by the segmentation CNN. 

%This strategy effectively assumes that any object is identified (segmented) on at least two images. 

\vskip 0.2cm     
{\bf MRF formulation}
We consider the space $\mathcal{X}$ of all {\it pairwise} intersections of view-rays from camera locations (c.f. Fig.~\ref{figIntersect}).
Any location $x\in \mathcal{X}$ is generated by the intersection of two rays $r_1$ and $r_2$ from camera view pair.
The  binary label $z\in \lbrace 0,1\rbrace$ is associated to $x$ to indicate the presence ($z=1$, referred to as positive intersection depicted as blue dots in Fig.~\ref{figIntersect}) or absence ($z=0$, empty intersection, yellow in Fig.~\ref{figIntersect}) of the object of interest at the corresponding intersection (see side panel in Fig.~\ref{figIntersect}). 
The space $\mathcal{Z}$ of all intersections' labels is then a binary  MRF~\cite{mrfs}, which is formally defined as follows:
%where objects are in locations with label value '1' (we will refer to these as positive intersections), and those with $z=0$ are empty intersections. 

$\bullet$  To each site $x$ we associate two Euclidean distances $d_{1}$ and $d_{2}$ from cameras: $d_{j}=\|C_{j}-x\|$, where $C_{j}$ are locations of two cameras $(j=\lbrace 1,2\rbrace)$ from which intersection $x$ is observed along the rays $r_{1}$ and $r_{2}$, respectively. Any intersection $x$ is considered in $\mathcal{X}$, only if $d_{1},d_{2}< 25\text{m}$. In Fig.~\ref{figIntersect}, red intersection in the upper part of the scene is rejected as too distant from Camera3.

$\bullet$ The neighborhood of node $x$ is defined as the set of all other locations $x_k$ in $\mathcal{X}$ on rays $r_{1}$ and $r_{2}$ that generate it. We define the MRF such that the state of each intersection depends only on its neighbors on the rays.
Note that the number of neighbors (i.e. neighborhood size) for each node $x$ in $\mathcal{X}$ in our MRF is unique.

$\bullet$ Any ray can have at most one positive intersection with rays from any particular camera, but several positive intersections with rays generated from different cameras are allowed, e.g. multiple intersections for Object1 in Fig.~\ref{figIntersect}.

\vskip 0.15cm
{\bf MRF energy}
The MRF configuration is defined by $\lbrace (x_i,z_i)\rbrace_{i=1,\cdots,N_{\mathcal{Z}}}$. 
For each site $x$ with state $z$ the associated MRF energy~\cite{mrfs} is composed of the following terms:
\vskip 0.1cm
$\bullet$ A unary energy term enforces consistency with the depth estimation. Specifically, the deep learning pipeline for depth estimation provides estimates $\Delta_1$ and $\Delta_2$ of distances between camera positions and the detected object at location $x$. We formulate the term as a penalty for mismatch between triangulated distances and depth estimates:
\begin{equation}
u_1(z)=z \sum_{j=1,2}\|\Delta_j-d_j\|
\end{equation}

$\bullet$ Pairwise energy term is introduced to penalize: ({\it i}) multiple objects of interest occluding each other, and ({\it ii}) excessive spread in case an object is characterized as several intersections.
In other words, we tolerate several positive intersections on the same ray only when they are in close proximity. This may occur in multi-view scenario due to segmentation inaccuracies and noise in camera geotag. For example, in Fig.~\ref{figIntersect} Object1 is detected as a triangle of positive intersections (blue dots) - two on each of the three rays. 

Two distant positive intersections on the same ray correspond to a scenario when an object closer to the camera occludes the second more distant object. Since we consider compact objects with negligible volume we can assume this type of occlusion unlikely.

This term depends on the current state $z$ and those of its neighbors $z_k$. It penalizes proportionally to the distance to any other positive intersections $x_k$ on rays $r_{1}$ and $r_{2}$:
\begin{equation}
u_2(z)= z \sum_{k} z_k \ \|x-x_k\|
\end{equation}

$\bullet$ A final energy term penalizes rays that have no positive intersections: false positives or objects discovered from a single camera position (see Fig.~\ref{figIntersect})\footnote{It is possible to register such rays as detected objects by applying the depth estimates directly to calculate the geotags. In this study we choose to discard such rays to increase robustness to segmentation false positives.}. This can be written as:
\begin{equation}
u_3(z)= (1-z) \prod_{k} (1-z_k)
\end{equation}
%where $k$ indexes all other intersections along the two rays defining current intersection.

%\vskip 0.1cm
The full energy of configuration ${\bf z}$ in $\mathcal{Z}$ is then defined as sum of energy contributions over all $N_{\mathcal{Z}}$ sites in $\mathcal{Z}$:
\begin{equation}
\mathcal{U}(\mathbf{z})=\sum_{i=1}^{N_{\mathcal{Z}}} \Bigl[ \alpha\  u_1(z_i)
+\beta\  u_2(z_i)+
(1-\alpha-\beta)\  u_3(z_i)\Bigr],
\label{EqEnergy}
\end{equation}
subject to $\alpha,\beta\geqslant0$, $\alpha+\beta\leqslant1$. 

\vskip 0.1cm
The optimal configuration is characterized by the global minimum of the energy $\mathcal{U}(\mathbf{z})$. The terms $u_1$ and $u_2$ penalize too many objects by increasing the total energy (for any positive intersection with $z=1$ both $u_1\geqslant0$ and $u_2\geqslant0$), whereas $u_3$ penalizes too few objects ($u_3=1$ for any ray with no positive intersections). The MRF formulation can also explicitly accommodate object location pattern assumptions through additional higher-order penalty terms.

\vskip 0.15cm
{\bf MRF optimization}
We perform energy minimization with Iterative Conditional Modes algorithm~\cite{mrfs} starting from an empty configuration: $z^0_i=0, \forall z_i \in \mathcal{Z}$. This (local) optimization is run according to a random node-revisiting schedule until local minimum is reached and no further changes are accepted. Experimentally, we have observed stable performance of the optimizer over multiple reruns. Therefore, there is no need to explore any more accurate global optimizers, see in~\cite{mrfs}.

\vskip 0.15cm
{\bf Clustering}
The final object configuration is obtained by resorting to clustering in order to identify groups of positive intersections that describe the same object\footnote{Operationally, clustering is also useful for parallelizing the geotagging step by splitting the analyzed large area in several smaller connective  parts with overlap - 25m buffer zone that belongs to areas on both sides of any border. This strategy allows us to retain all rays for triangulation while clustering resolves the object redundancies when merging the parts.}. Indeed this is required since we consider the space $\mathcal{X}$ of all pairwise intersections, whereas some objects are observed from three or more camera positions and result in multiple detected object instances. For example, in Fig.~\ref{figIntersect} Object1 is identified as three distinct positive intersections tightly scattered around the object. In this work we employ hierarchical clustering with a maximum intra-cluster distance of 1m which corresponds to our object sparsity assumption. Intersection coordinates for locations in the same cluster are averaged to obtain the final object's geotag.

\section{Experimental study}
\label{sec:exps}

%In this section we present results obtained on Google Street View imagery. 
We test our pipeline on GSV imagery for detection of two object types:  traffic lights and telegraph poles. Both kinds of objects are compact enough to be attributed a single geotag and are predominantly visible from roads. Traffic lights are  visible by design and telegraph poles are also typically erected in the close vicinity of the road network. 

%Both object-classes are appropriate for detection from street-side imagery: whereas the former are necessarily visible from roads by design, the latter are typically found along the existing roads and the majority is visible in the close vicinity of road network. Both types of objects have relatively small volume and can be attributed a single geotag. 

\begin{figure}[t!]
\centering\lineskip=-9pt
Correct\qquad\qquad\quad Partial\qquad\quad\qquad False positives\\
\vskip 0.12cm
\subfigure{\includegraphics[width = 0.325\linewidth]{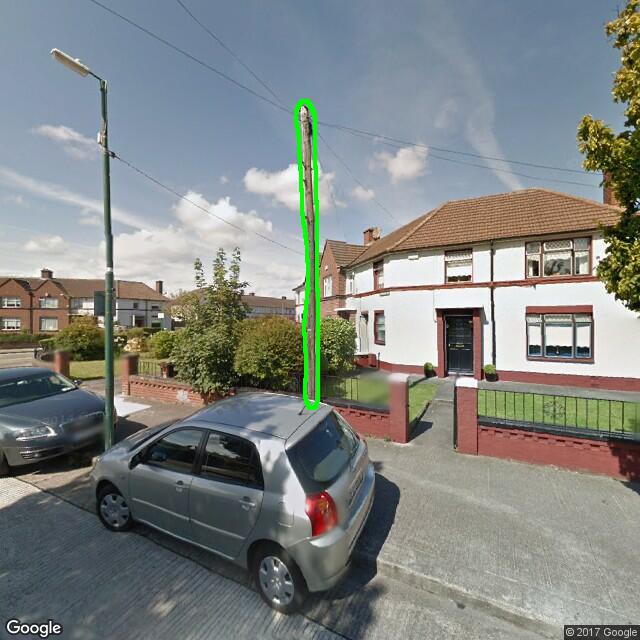}}
\subfigure{\includegraphics[width = 0.325\linewidth]{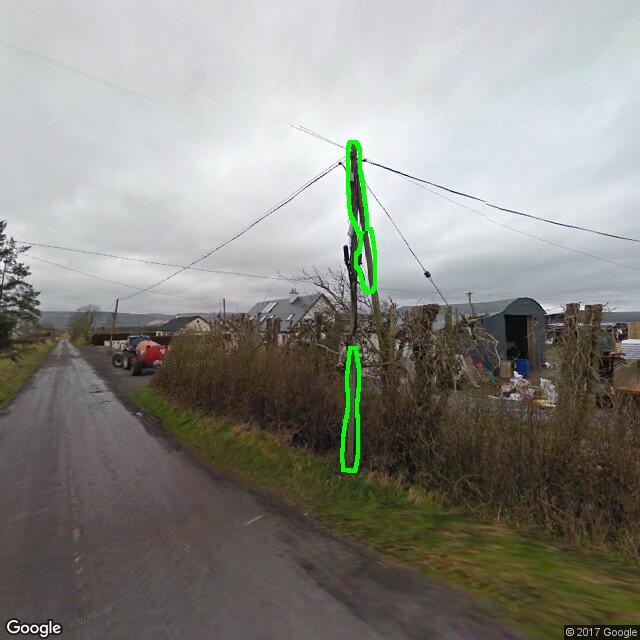}}
\subfigure{\includegraphics[width = 0.325\linewidth]{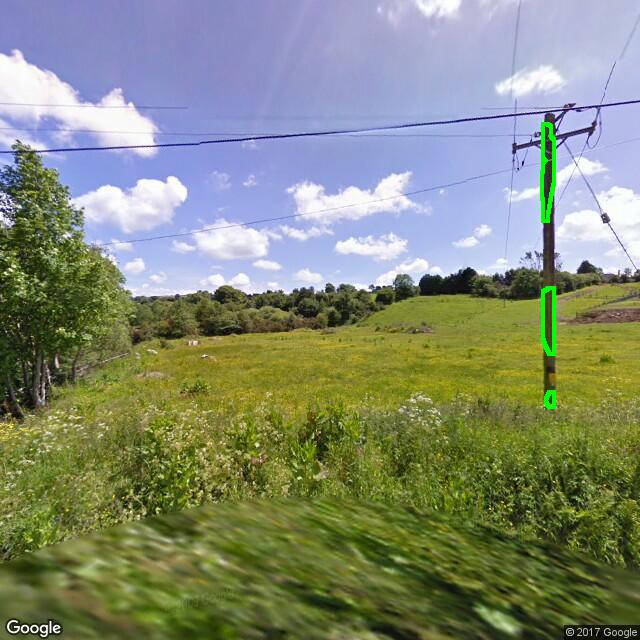}}\\
\subfigure{\includegraphics[width = 0.325\linewidth]{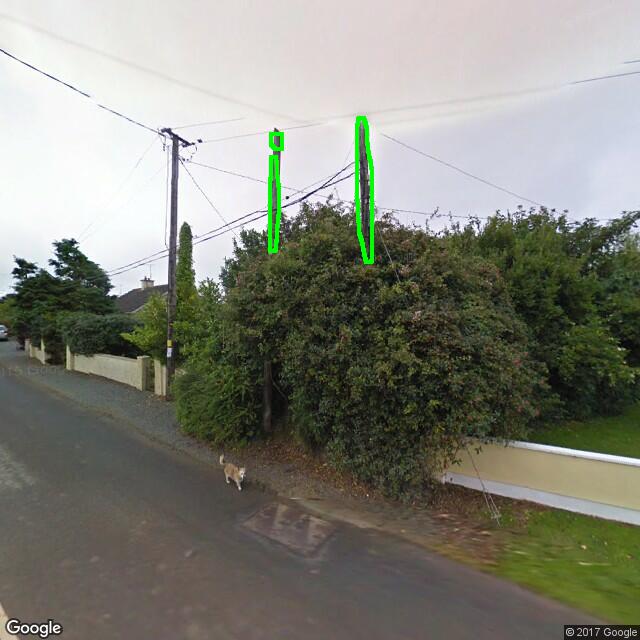}}
\subfigure{\includegraphics[width = 0.325\linewidth]{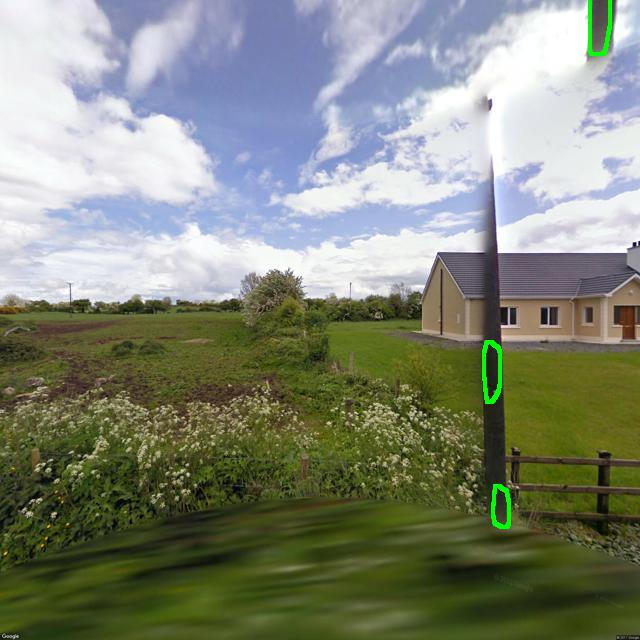}}
\subfigure{\includegraphics[width = 0.325\linewidth]{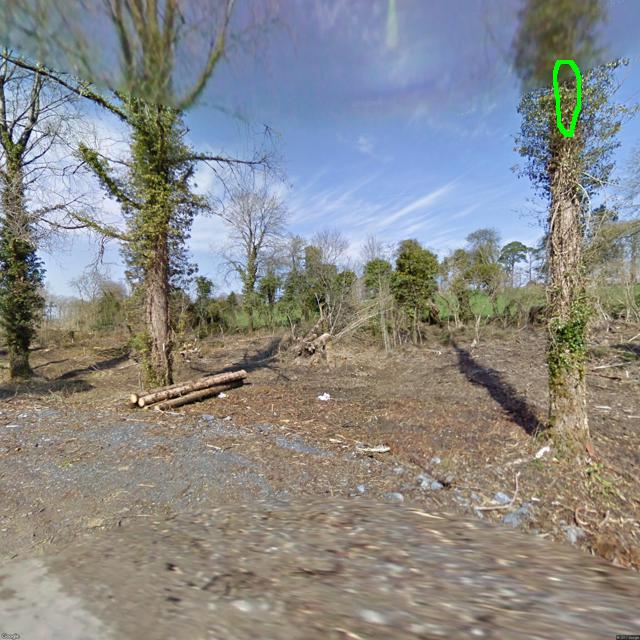}}
\caption{Telegraph pole segmentation on Google Street View. First row: correct segmentation. Second row: partial segmentation. Third row: false positives (electricity pole, ivy-covered tree).}
%Telegraph pole segmentation on Google Street View images. First two rows: correct segmentation. Third row--- partial detection, fourth row--- partial detection and strong stitching, fifth row--- false positives: ivy-covered tree and electricity pole.}
\label{FigPoleSeg}
\end{figure}

In order to evaluate the accuracy of the estimated GPS coordinates we deploy our detector in several areas covered by GSV imagery. To extract the latter we first create a dense grid of GPS coordinates along road-centers with 5m step and then query the API to retrieve closest available GSV panoramas in 4 parts each with 90 degrees field of view at $640\times640$ pixels resolution. Note that GSV imagery often demonstrates strong stitching artifacts, in particular in the top and bottom parts of the images, see examples in Fig.~\ref{FigPoleSeg}.

For the segmentation FCNN we use a {Caffe} implementation and perform training on a single NVIDIA TITAN X GPU using stochastic gradient descent with fixed learning rate of 5e-11 and momentum $m=0.99$, with batches of 2 images. This choice is in line with recommendations in~\cite{shelhamer2017fully} and empirically demonstrated the best performance. We resort to data augmentation by performing random horizontal flipping, rotations $[-5^{\circ},5^{\circ}]$, small enhancements of input image's brightness, sharpness and color. 
The imbalance between classes - object and background - in the datasets is handled well by the FCNN provided we maintain a fraction of object-containing images in the training set above $25\%$ at all times. The inference speed is at 8 {\it fps}. The depth estimation pipeline~\cite{laina2016deeper} is used in the authors' implementation in {MatConvNet} with no modifications.
The energy weights in Eq.~(\ref{EqEnergy}) are set to $\alpha = 0.2$ and $\beta=0.6$, both segmentation FCNN loss functions have the same weight.

%The employed FCNN pipelines report good results overall (see more detail below). Together with the MRF formulation these allow us to accurately recover the vast majority of objects from the employed imagery. Hence, in the following we focus on evaluating the performance of the geotagging module, which is central to this work, and consider outside of scope any further comparative analysis with alternative segmentation and depth estimation techniques, e.g.~\cite{liu2016ssd,GodardAB16}.

%%%%

\subsection{Traffic lights}
\label{sec:exp:traffic}

%\subsubsection{FCNN  training for Segmentation}
The image segmentation pipeline for traffic lights is trained on data from two publicly-available datasets with pixel-level annotations: Mapillary Vistas~\cite{mapillary} and Cityscapes~\cite{cityscapes}. We crop/resize the images to match the standard GSV image size of $640\times640$. This provides us with approx. 18.5K training images containing large traffic lights (at least $10\times10$ pixels each instance).  We start with a  PASCAL VOC-pretrained model and carry on training with learning rate of 5e-11 for another 200 epochs.

\begin{figure}[t!]
\centering\lineskip=-9pt
Correct\qquad\qquad\quad Partial\qquad\quad\qquad False positives\\
\vskip 0.12cm
\subfigure{\includegraphics[width = 0.325\linewidth]{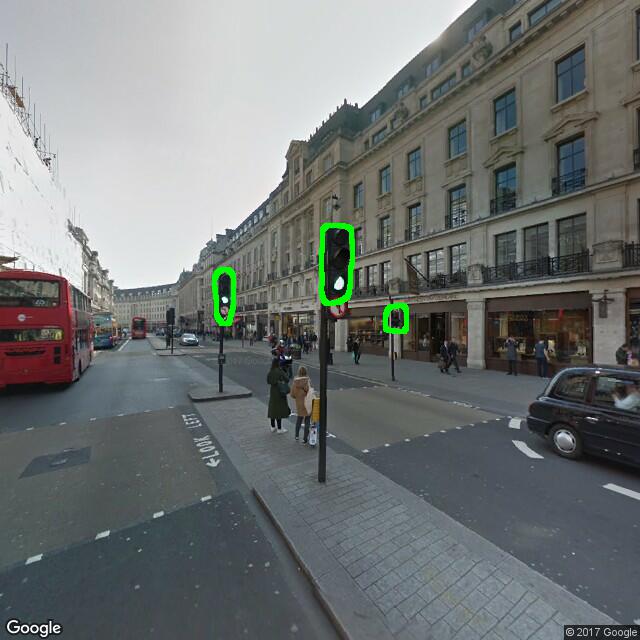}}
\subfigure{\includegraphics[width = 0.325\linewidth]{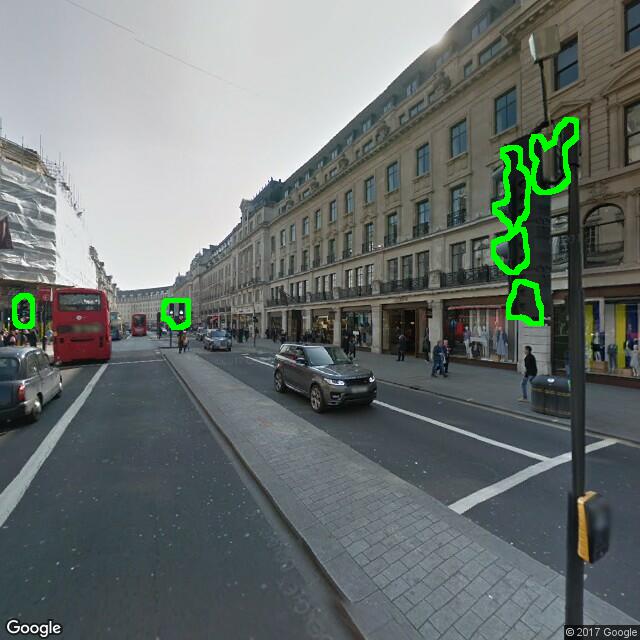}}
\subfigure{\includegraphics[width = 0.325\linewidth]{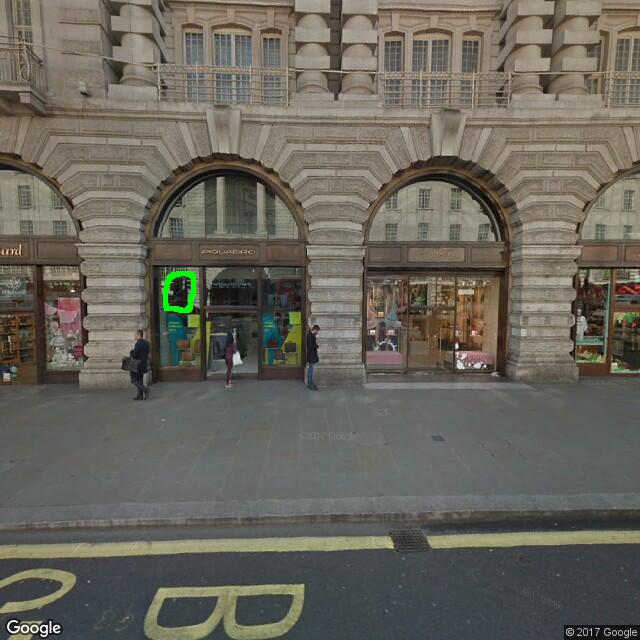}}\\
\subfigure{\includegraphics[width = 0.325\linewidth]{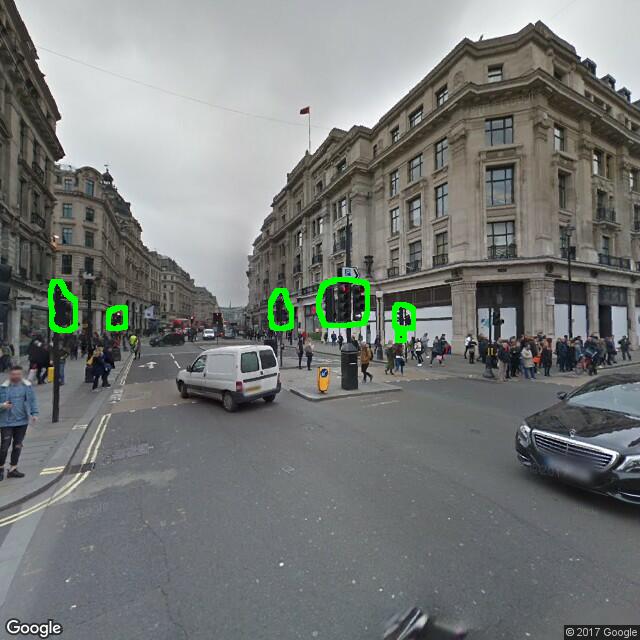}}
\subfigure{\includegraphics[width = 0.325\linewidth]{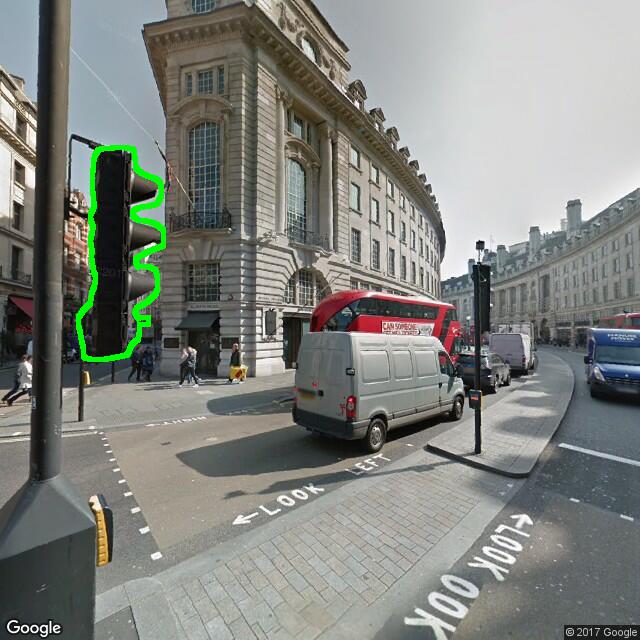}}
\subfigure{\includegraphics[width = 0.325\linewidth]{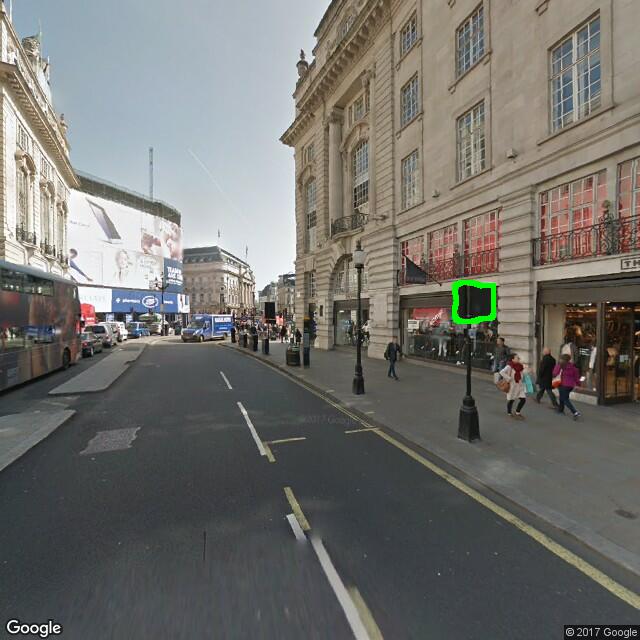}}
\caption{
%Traffic light segmentation on Google Street View images. First two rows: correct segmentation. Third row--- noisy segmentation, left image fourth row --- false negative (traffic light from rear). Rest--- false positives: reflected traffic light, other objects.}
Traffic light segmentation with Google Street View. First column: correct segmentation. Second column: partial segmentation. Third column: false positives (reflection, traffic sign).}
\label{FigTLSeg}
\end{figure}

{\bf Object detection}
The test pixel-level precision plateaus at $.723$ and recall at $.735$. If over $25\%$ of an object's pixels are labeled correctly we consider such object instance to be recovered accurately. In the vicinity of camera (within 25m from camera position) the instance-level precision is $.951$ and recall is $.981$ on the test set. 
In this experiment the introduction of the second FCNN loss function has improved the instance-level precision by $.022$ and decreased the recall by $.002$.
In general, instance-level recall is a more important characteristic for geotagging of objects, whereas a lower precision (false positives) is partially compensated at the later stage by the employed tagging procedure which requires objects to be observed {\it multiple} times. In this study we adopt a conservative strategy of ignoring objects located farther than 25m from camera positions due to the pronounced performance drop of image processing reliability on distant objects: substantial decrease of recall rate in semantic segmentation and high variance in depth estimation. Note that farther objects may be discovered through semantic segmentation but are rejected by the geotagging procedure described in Sec.~\ref{sec:GPStagging} through the maximal distance restriction.

In Fig.~\ref{FigTLSeg} we show examples of traffic light segmentation in busy urban scenes. The performance is overall very good with minor issues of occasional small false positives and on some traffic lights facing away from the camera (see images in central and right columns). Neither of the two poses particular problems for geotagging since the processing is done on series of images, which allows us to observe most traffic lights both from front and rear and discard false positives due to their inconsistent and / or non-stationary nature. As far as the pixel-level performance is concerned a high precision rate is not necessary since we require pixel-level labels solely to access the relevant depth estimates obtained through the depth evaluation pipeline.

\begin{figure}[t!]
\begin{center}
\includegraphics[width = \linewidth]{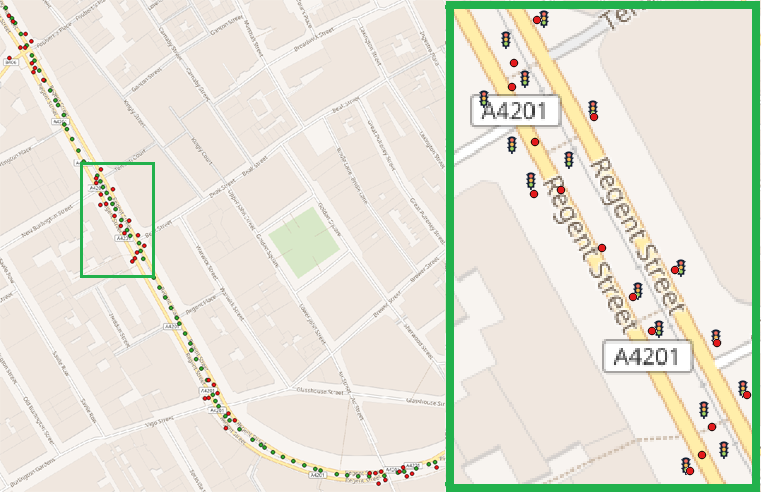}
\end{center}
\vspace{-0.5\baselineskip}
\caption{Map of traffic lights automatically detected in Regent Street, London, UK. Green dots, red dots and traffic lights symbols (in zoom) are the GSV camera locations, geotagging results and the actual locations of objects, respectively.}
\label{figRegent}
\end{figure}

\begin{table}[!b]
\tabcolsep = 0.75mm
\centering
\begin{tabular}{|c||c|c|c|c|c|c|c|}
\hline
 & \#Actual & \#Detected & TP & FP & FN& Recall & Precision\\
\hline\hline
\includegraphics[width = 0.45cm]{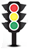} & 50 & 51 & 47 & 4 & 3 & .940 & .922\\
\hline
\includegraphics[width = 0.45cm]{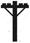} & 77 & 75 & 72 & 3 & 5 & .935 & .960\\
\hline
\includegraphics[width = 0.25cm]{images/pl.png}\includegraphics[width = 0.25cm]{images/pl.png} & 
{2696} & 
{2565} & 
{2497} & 
{68} & 
{199} & 
{.926} & 
{.973}\\
\hline
\end{tabular}
\caption{Instance-level object discovery results for traffic lights and telegraph poles in the test areas. (TP: True Positives, FP: False Positives, FN: False Negatives)} 
\label{tab:confusion:matrices}
\end{table}

% 5 clusters = 6+8+16+13+7
{\bf Geotagging}
To evaluate the geotagging performance we consider a 0.8 km stretch of Regent street in London, UK, covered by 87 GSV panoramas, see Fig.~\ref{figRegent}. This dense urban area has five clusters of traffic lights, totaling 50 individual objects. To avoid ambiguity in the count we assume that traffic lights with multiple sections (e.g., separate sections for cars and pedestrians) attached to a single pole are counted as a single object. 
%The configuration of objects in this area satisfies the sparsity assumption. 
Note that we do not have access to precise GPS coordinates of traffic lights and perform our analysis based on a human interpretation of GSV images.

Segmentation reports 179 single-view traffic lights instances with 70 objects in 51 clusters after geotagging. We consider objects to be recovered accurately if they are located within 2 meters from the reference position. This choice is in line with the official reports on the accuracy of GPS measurements that establish a $1.89$m $95\%$-confidence interval for horizontal error for single-frequency GPS receivers~\cite{gpsfaa}. {The object discovery results are summarized in Table \ref{tab:confusion:matrices}. The obtained object recovery precision and recall are both above $.92$ and consistently outperform results reported for telecom assets recognition in~\cite{wacv17} with less than $.85$.}
Of the 4 false positives 2 correspond to actual objects that are approx. 3m from reference locations. 6 single-view instances were not matched: 4 segmentation false positives and 2 objects identified from one view.

%reports 47 true positives 4 false positives and 3 false negatives, hence, recall of $94\%$, precision --- $92\%$. Out of the 4 false positives 2 correspond to actual objects, but are at roughly 2.5m distance from ground truth locations. Results are summarized in Table \ref{tab:confusion:matrices}.

%%%%%%%%%%%%%%%

\subsection{Telegraph poles}
\label{sec:exp:poles}
%\subsubsection{FCNN training for Segmentation }
Segmentation of the telegraph poles relies on a custom training dataset. Indeed, even though multiple existing datasets incorporate class `poles', none of them provides sufficient distinction between different types of poles: lampposts, poles carrying traffic lights and / or road signs, bollards (sidewalk poles), and utility poles (electricity, telegraph). We consider here a particular class of telegraph poles that are actively used in the Republic of Ireland. The corresponding network is very large with over a million of poles spread throughout the country. Overall, the main challenge in detection of such poles is their strong visual similarity to other tall poles, in particular, electricity poles. 

%All of these poles are made of wood and visually have a natural `wooden' texture (no artificial coloring due to paint). Telegraph poles are not used for mounting lamps and there is no shared use of poles with electricity distribution network. Hence, this typology of poles is a clearly defined subset of all {\it tall} poles, that include all lampposts and utility poles. There are several more subtle distinctive visual features about telegraph poles, like steps for climbing, particular types of insulation in use, and specific types of objects mounted on the poles, but none are necessarily present (or visible) on any particular pole. Telegraph poles are occasionally covered by vegetation: overgrown by ivy or closely surrounded by trees. 

\begin{figure}[t!]
\begin{center}
\includegraphics[width = \linewidth]{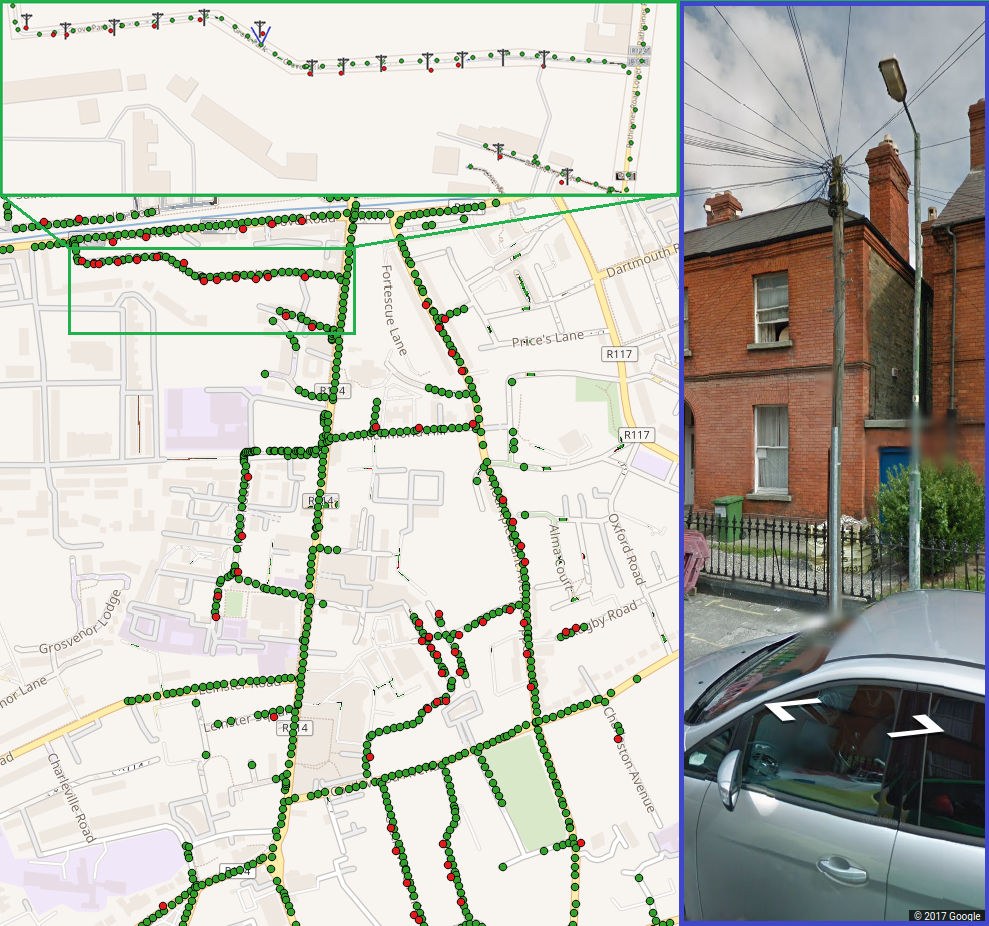}
\end{center}
\vspace{-0.5\baselineskip}
\caption{Map of telegraph poles detected in Dublin, Ireland. 
Green dots, red dots and pole symbols (in zoom) are the GSV camera locations, detected and actual locations of objects, respectively. The image in blue is a GSV image of one of poles with image position and geo-orientation set up based on our geotag estimate.}
\label{figRath}
\end{figure}

To prepare a segmentation pipeline for this object class we adopt the following procedure: we first train our FCNN to detect {\it all} tall poles: utilities and lampposts, and then fine-tune it to extract only telegraph poles. We do the former by combining existing datasets,  Mapillary Vistas~\cite{mapillary} and Cityscapes~\cite{cityscapes}, with GSV imagery extracted at locations provided through a training GPS database of telegraph poles made available to us for this study. Note that this database cannot be used directly for segmentation training for the following reasons: First, because of the inherent GPS inaccuracy in both pole and GSV coordinates, and frequent occlusions, one cannot expect to observe poles in the center of images extracted based on geotags alone. Second, many images depict geometry rich scenes, with mixtures of different pole-types and occasional other strong vertical features, such as trees, facade elements, antennas, etc. To train the FCNN to distinguish tall poles from any other objects in the scenes we put together a dataset with 20K annotated images. We run this first step of training for 100 epochs to achieve satisfactory discriminative power. As a second step we fine-tune this network on our custom pixel-level annotated set of $500$ telegraph pole images. To further boost the discriminative power of the FCNN we add 15K GSV scenes collected in areas with no telegraph poles but in presence of lampposts and electricity poles. The training is run for another 200 epochs at the learning rate of 2e-11.

{\bf Object detection}
Sample pole segmentation maps are presented in Fig.~\ref{FigPoleSeg}.
The final FCNN reports pixel-level recall of $.936$ and precision $.822$. As above, we consider relevant only the objects within 25m of the camera. This ensures accurate distinction from other tall poles since the decisive subtle differences are not visible at greater distances. The object-level test recall is $.979$ and precision is $.927$ .

We perform experimental analysis in the Rathmines area of Dublin, Ireland, see Fig.~\ref{figRath}. The area under study has approx. 8 km of roads covered by 945 GSV panoramas. 77 telegraph poles are visible  through GSV imagery. 273 single-view instances are discovered via image segmentation and the geotagging results are summarized in Table~\ref{tab:confusion:matrices}. Out of the 3 false positives: 2 are wrong objects and 1 is a telegraph pole 3m away from its true location.

{\bf Geotagging}
We now evaluate the accuracy of geotags compared to the ground truth manually collected with a single-frequency GPS device: each pair of coordinates has been recorded three times on three different days and averaged to produce a single reference geotag for each object. This ground truth contains 41 poles in the considered area. These poles have been automatically detected in 89 GSV images, and all 41 discovered in at least 2 images enabling our method to infer their GPS coordinates.
%36 poles are seen by  two images views, 3 poles are seen on three images, and 2--- on four. 
%The 41 poles with ground truth GPS coordinates are observed on 89 images: 36 objects are recovered on two images, 3--- on three images, and 2--- on four. 

{To compare the accuracy of mapping with~\cite{wacv17}, we have estimated the  relative distance precision (calculated distance / ground truth object-to-camera distance * $100\%$). With this metric we achieve $95.8\%$ average and $18\%$ standard deviation which improves on $70\%$ average reported in~\cite{wacv17}. Additionally, we evaluate the absolute accuracy of GPS positioning of objects: Table \ref{tab:gps:accuracy} reports the empirical statistics of object position estimation error (absolute value of the distance between the reference and triangulated positions) : mean, median, variance, $95\%$-empirical confidence interval. Notably the obtained interval of $2.07$m is comparable with the $1.89$m confidence interval for single-frequency GPS receivers~\cite{gpsfaa}. Practically, this implies that the accuracy of automatic mapping achieved by our method is equivalent to that of a human operator manually performing object mapping with a commercial-level GPS-enabled device.}

\begin{table}[!t]
\tabcolsep = 1.3mm
\begin{tabular}{l|cccc}
method &mean&median& variance & 95\% e.c.i. \\
\hline
MRF-triangulation & 0.98&1.04 &0.65 &2.07 \\
%GT & - & - & - & 1.89$^{*}$ \\
depth FCNN & 3.2&2.9 &2.1 &6.8 \\
\hline
\end{tabular}
\caption{Accuracy of geotagging. Distance statistics in meters from reference data collected on site with a GPS receiver}
\label{tab:gps:accuracy}
\end{table}

In Table~\ref{tab:gps:accuracy} we report the results of geotagging based on depth estimation from 89 monocular views without resorting to triangulation.
%analysis of depth estimation accuracy for telegraph poles. 
%The comparison of 89 true values of camera-pole distance (from ground truth) with depth estimates (obtained via FCNN pipeline) results in the following values for bias (absolute difference between distance and depth): mean of 3.2m, median 2.9m, variance 2.1m, and $95\%$-empirical confidence interval of 6.8m. 
We observe a substantial improvement of GPS accuracy obtained with our triangulation
procedure.
In both cases - MRF-triangulation and depth FCNN - the geotags rely entirely on the accuracy of the input camera coordinates provided with GSV imagery. Any outliers in camera geolocations result directly in object geotag errors.

{\bf Large scale study} {To validate the performance on a large-scale problem we have deployed the detector on a cluster of 120 kms of public roads in County Mayo, Ireland, captured by $13500$ GSV panoramas. This area has extensive networks of telegraph and electrical poles and we had the ground truth for $2696$ telegraph pole positions. Since the GPS-accuracy of ground truth records is not homogeneous we utilize these solely to validate the recall / precision of the pipeline and not the GPS-tagging accuracy. A detected pole is considered true positive (TP) if it is within 3 meters of a recorded object. The validation results are summarized in Table~\ref{tab:confusion:matrices} (bottom row). We observe similar performance as reported above. A slightly lower recall occurs because the ground truth records are collected manually and include many road-side poles completely occluded from view by vegetation. Higher precision is achieved due to the relative complexity of the urban environment compared to the mostly rural imagery depicting less challenging geometries processed in this experiment.\\
In this experiment an average of 1.56 objects is typically observed per image reporting any detections. This occurs because telegraph poles are typically parts of longer pole-lines and several identical poles are often visible from the same camera position. This highlights the necessity for non-standard triangulation such as our MRF-based technique to enable object mapping. In the considered road cluster around $35\%$ of the pole-containing scenes cannot be mapped using methods similar to those proposed in~\cite{wacv17,soheilian2013detection,iccv11} due to the presence of multiple visually identical telegraph poles, which renders visual matching-based triangulation inefficient.}

\section{Conclusions}
\label{sec:concl}
We have proposed an automatic technique for detection and geotagging of recurring stationary objects from street view imagery. 
The proposed solution relies on two existing deep learning pipelines, one fine-tuned for our needs while the other is employed as-is. A novel triangulation-based MRF has been formulated to estimate object geolocations that allows us to handle automatically any multi-object scenes. The triangulation helps to avoid duplicates and simultaneously reduce the number of false positives. This geotagging module is independent and can be combined with any other segmentation and / or depth estimation pipelines fine-tuned for specific object classes. The experimental analysis has demonstrated high object recall rates. 

The empirical accuracy of the geotagging is within 2 meters, similar to that obtained with a single-frequency GPS receiver. {The main source of GPS-inaccuracies is noise in the camera position coordinates. In other words, further improvement of mapping accuracy should be addressed by suppression of camera position noise or outliers.}

%The proposed solution is based on two FCNNs - one for semantic labeling, another for monocular image depth estimation, and a novel triangulation-driven optimization procedure for recovering geotags. We have trained the segmentation pipeline according to the two objects classes considered: traffic lights and telegraph poles. The experimental analysis has demonstrated high accuracy performance in both discovering objects and geotagging.

One of the directions for future work would be the design of a unified FCNN architecture for segmentation and depth which, in the absence of a priori depth data, can be trained as the second stage on the results of the triangulation procedure presented here. This may allow partial or complete relaxation of the sparsity assumptions by providing improved input to the triangulation-based geotagging pipeline.

\section*{Acknowledgement}
This research was supported by {\it eir} - the principal provider of fixed-line and mobile telecommunications services in Ireland - who are actively investing in research into machine learning and image processing for the purposes of network planning and maintenance.
%We are particularly grateful to Sean Abraham and Frank Walsh who provided expertise in the preparation of the custom dataset for poles and feedback on initial outputs to help fine tune the accuracy of the FCCN.

{\small
\bibliographystyle{ieee}
\bibliography{biblio}
}   

\newpage
\clearpage
\section*{Supplementary material}

\subsection*{Panoramic stitching}
The employed GSV imagery demonstrates strong stitching artifacts. In particular, the horizontal stitching is always present in the top and bottom parts of the images (queried at horizontal orientation with tilt = 0), see Fig.~\ref{FigTLSeg}, Fig.~\ref{FigPoleSeg} and Fig.~\ref{FigPoleSeg2}. If a vertical object crosses the top stitching line, its upper part is disconnected from the rest of the object. The geo-orientation of the objects (clockwise rotation of the camera from north to the centre of the current scene) cannot be correctly inferred from these disconnected upper parts. To avoid incorrect geotagging we mask the output of the segmentation pipeline to remove detection above the top stitching line.

\subsection*{2-stage telegraph pole segmentation training}
The challenge of telegraph pole segmentation is in the absence of a publicly available dataset that would allow enough training to accurately distinguish this particular class of poles from other poles and similar vertical objects, in particular electricity poles, lampposts, trees with little or no vegetation. The considered telegraph poles are made of wood and visually have a natural `wooden' texture (no artificial coloring due to paint). They are never used for mounting lamps and there is no shared use of poles with electricity distribution network. Hence, this typology of poles is a clearly defined subset of all {\it tall} poles, that include all lampposts and utility poles. There are several more subtle distinctive visual features about telegraph poles, like steps for climbing, particular types of insulation in use, and specific types of objects mounted on the poles, but none are necessarily present (or visible) on any particular pole. 
Telegraph poles are occasionally covered by vegetation: overgrown by ivy or closely surrounded by trees.  

%Strong stitching often does not allow us to accurately identify the pole. This occurs when the camera is located very close to the pole and its upper half (containing visual information necessary to identify its type) is stitched incorrectly so that the pole is cut into two disconnected parts.

To train the segmentation FCNN we rely on a 2-stage training procedure: first, using the proprietary telegraph-pole GPS dataset and publicly available Cityscapes and Mapillary Vistas datasets to achieve detection of all poles, and, second, on our custom dataset with 500 fine pixel-level segmentations of telegraph poles and multiple negative examples to train the final segmentation pipeline. The first step is required since the 500-images custom dataset is not sufficient to train the FCNN to discriminate strong vertical features such as facade elements, fences, trees, etc.

The fixed learning rates are 5e-11 (100 epochs) and 2e-11 (200 epochs) for the first and second stages, respectively. These low learning rates are as suggested by {\it Shelhamer et al.}~[17] and allow the employed architecture (FCN-8s all-at-once) with deconvolutions and skip connections to be trained in an end-to-end fashion. Experiments with other (higher) learning rates and alternative optimizers (ADAM, ADAGRAD) reported worse convergences.

\begin{figure}[t!]
\centering\lineskip=-9pt
\subfigure{\includegraphics[width = 0.325\linewidth]{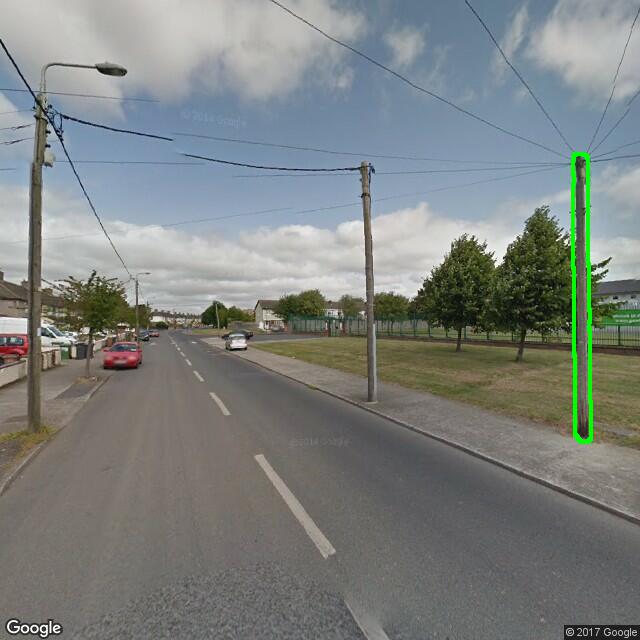}}
\subfigure{\includegraphics[width = 0.325\linewidth]{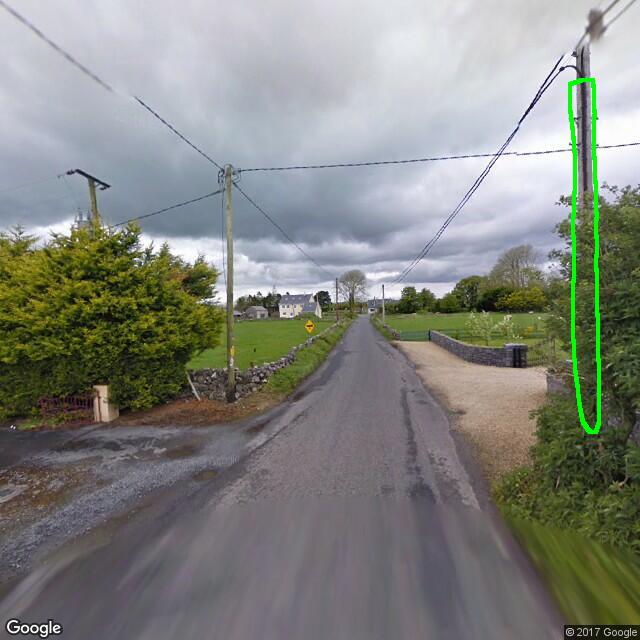}}
\subfigure{\includegraphics[width = 0.325\linewidth]{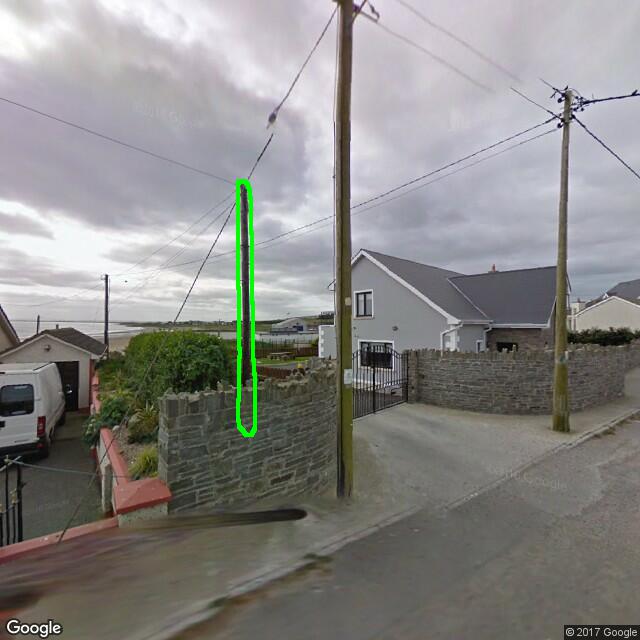}}\\
\subfigure{\includegraphics[width = 0.325\linewidth]{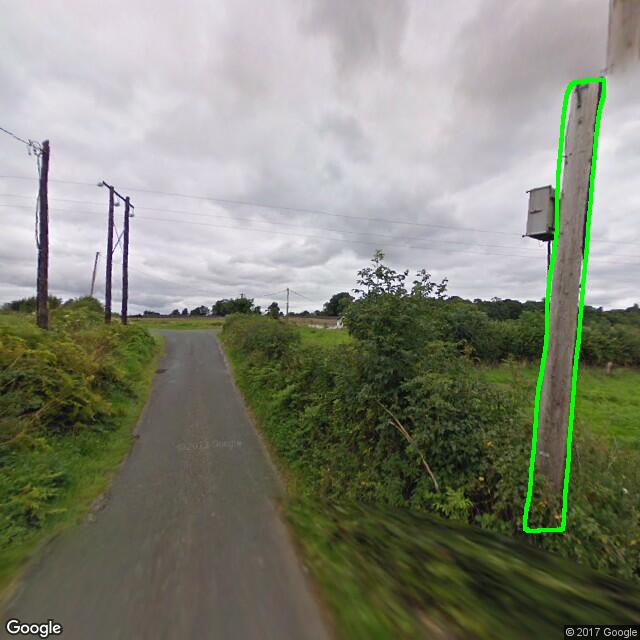}}
\subfigure{\includegraphics[width = 0.325\linewidth]{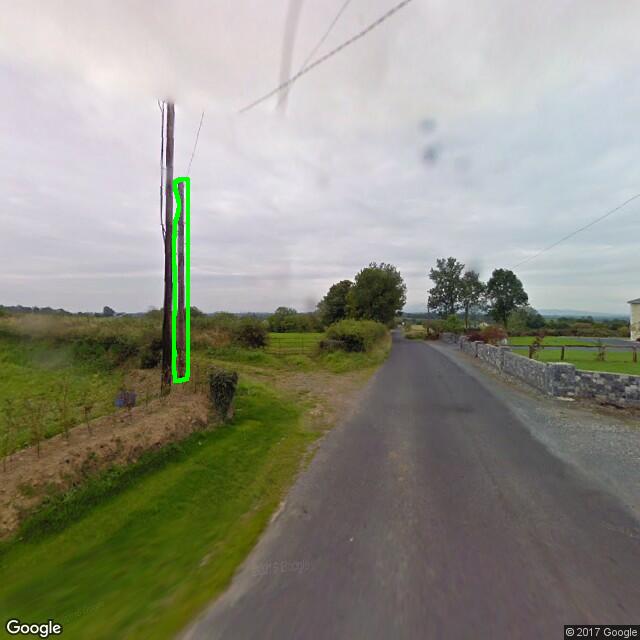}}
\subfigure{\includegraphics[width = 0.325\linewidth]{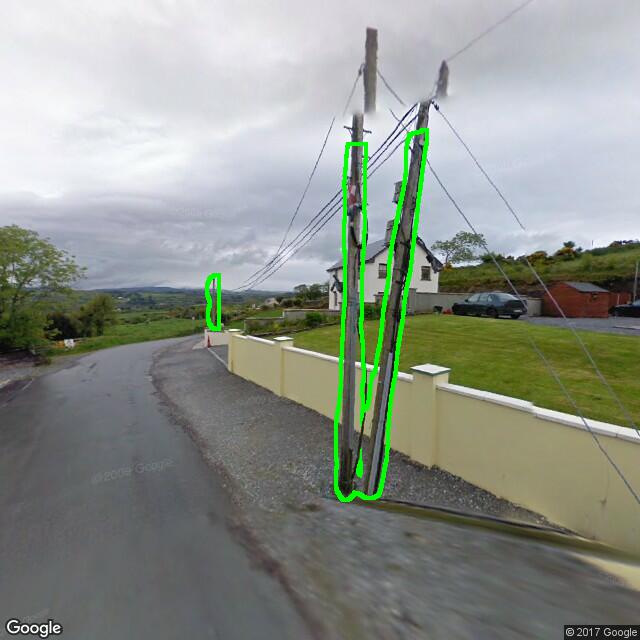}}\\
\subfigure{\includegraphics[width = 0.325\linewidth]{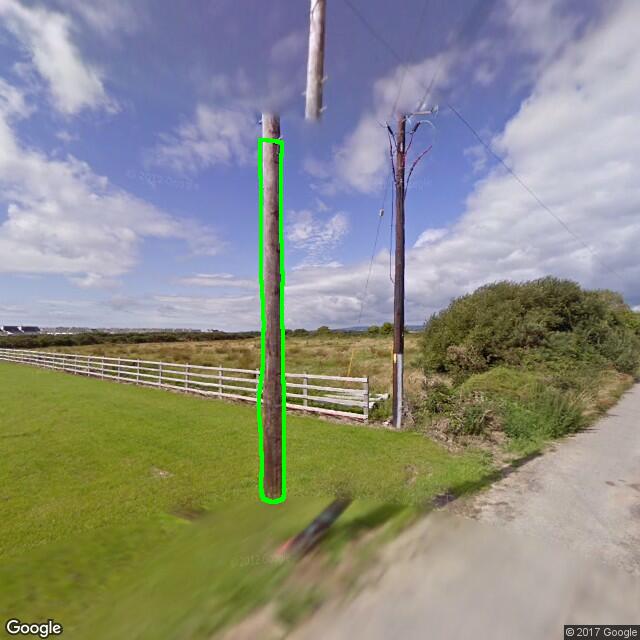}}
\subfigure{\includegraphics[width = 0.325\linewidth]{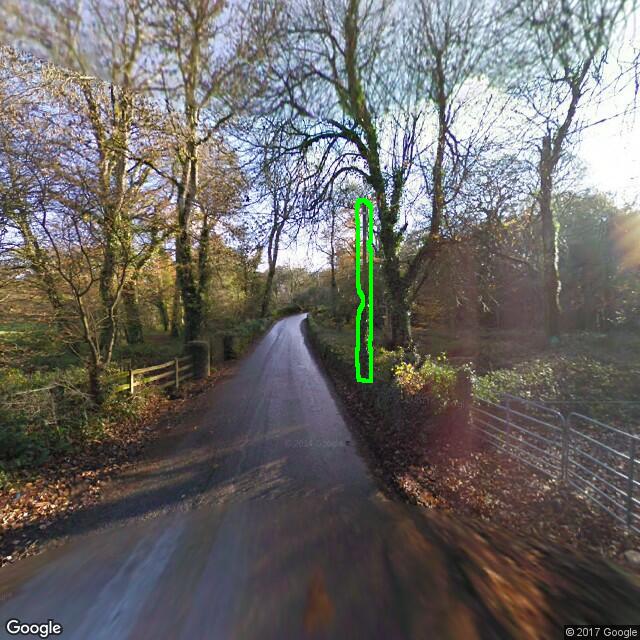}}
\subfigure{\includegraphics[width = 0.325\linewidth]{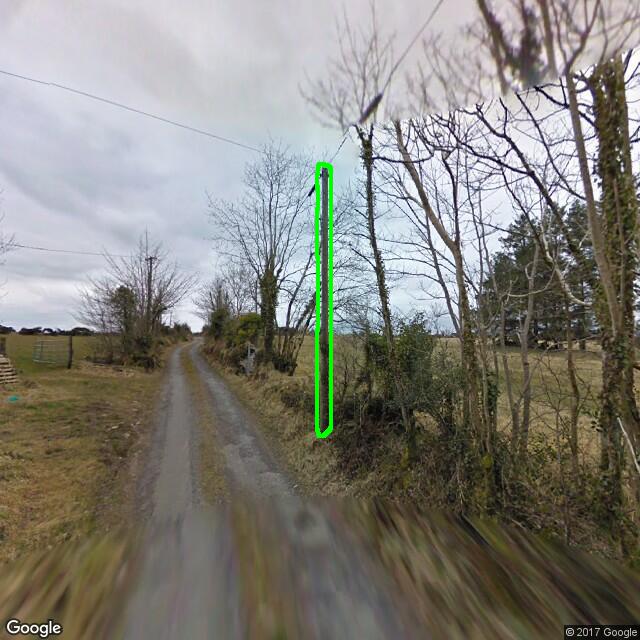}}
\caption{Examples of correct telegraph pole segmentation in the presence of other similar vertical objects, such as electricity poles, lampposts and trees.}
\label{FigPoleSeg2}
\end{figure}

In Fig.~\ref{FigPoleSeg2} we present multiple complex GSV scenes where our 2-step trained segmentation FCNN reported accurate results. All the rejected objects in these scenes are not telegraph poles but rather electricity poles, lampposts or trees.

\subsection*{Video}
A video demonstrating the performance of the telegraph pole detection on a sequence of 150 frames is available at \url{https://youtu.be/X0tM_iSRJMw}. For visualization we present the segmentation in the frontal 90-degrees field of view. The object detection and geotagging are performed based on the full 360-degree panoramas. For this reason most of the telegraph poles missed in the frontal parts of panoramas are discovered since the frontal frames represent 25\% of the total image coverage. 37 of 38 telegraph poles along the analysed stretch of road are successfully discovered. All the segmentation false positives are rejected automatically by the geotagging procedure and are not present in the final object list which is demonstrated on the right side of the video. Note that there are multiple electricity pole and trees in this area.

\end{document}